\documentclass[12pt]{article}
\usepackage{graphicx}

\begin{document}
\def\be{\begin{equation}}
\def\ee{\end{equation}}

%\begin{center}
%{Surajit Chakrabarti\\ Ramakrishnamission Vidyamandira, Belur, West Bengal\\}
%\end{center}

%\begin{center}
%{Soumen Sarkar$^1$, Sanjoy Kumar Pal$^2$, Surajit Chakrabarti$^3$ \\
%$^1$Karui P.C.High School, Hooghly, WB, India\\
%$^2$Anandapur H.S. School, Anandapur,Paschim Medinipur,WB,India\\
%$^3$Ramakrishna Mission Vidyamandira, Belur Math, Howrah, WB,India\\}
%E-mail: tosoumen84@gmail.com; sanjoypal83@gmail.com; surnamchakrabarti@gmail.com
%\end{center}

\begin{center}
{\large\bf{ Use of a smartphone camera to determine  the focal length of a thin lens by finding the transverse magnification  of the virtual image of an object}}
\end{center}
%\begin{center}
%{Surajit Chakrabarti\\ Ramakrishnamission Vidyamandira, Belur, West Bengal\\}
%\end{center}

\begin{center}
{Sanjoy Kumar Pal$^1$, Soumen Sarkar$^2$, Surajit Chakrabarti$^3$ \\
$^1$Anandapur H.S. School, Anandapur, Paschim Medinipur, WB, India\\
$^2$Karui P.C.High School, Hooghly, WB, India\\
$^3$Ramakrishna Mission Vidyamandira, Belur Math, Howrah, WB, India\\}
E-mail: sanjoypal83@gmail.com; tosoumen84@gmail.com; surnamchakrabarti@gmail.com
\end{center}
%\abstract
\begin{center}
{\large\bf{ABSTRACT}}
\end{center}

In this work we have determined the focal length of a concave lens by photographing the virtual image of an object by a smartphone camera. We have similarly determined the focal length of a convex lens by forming a virtual image of an object keeping it within the focal distance from the lens. When a photograph is taken by a smartphone, the transverse width  of the image on the sensor of the camera in pixels can be read off by software available freely from the internet. 
 By taking a photograph of the virtual image from two positions of the camera separated by a distance along the line of sight of the camera, we have determined the transverse width of the virtual image. From this we find the focal lengths of the lenses knowing the transverse width and the distance of the object from the lenses.
\newpage
\section{Introduction}
A smartphone has become a very useful tool for conducting physics experiments using the sensors in the phones. Many experiments have been done recently with a smartphone as can be found from the literature [1]. In the present work we have determined the focal lengths of a concave and a convex lens by finding the widths of the virtual images by photographing them with smartphone cameras.
A convex lens produces a virtual image of an object when it is placed within the focal distance from the lens. A concave lens always produces a virtual image of a real object.
Experimental determination of the focal length of a lens from the position of the virtual image is not common, for the simple reason that this image cannot be cast on a screen. However, by photographing the virtual image with a smartphone, this is possible. In a recent work [2] we have shown how to determine the lateral width of an object by taking its photograph from a distance. We will follow the principle of this work to find the width of the virtual image produced by a thin lens. In the  works [3,4] the method for determining the focal length of a smartphone camera has been discussed. In another work [5] the method for determining the magnification of a virtual image from the photograph by a smartphone has been discussed, though this work does not give any experimental data. Another work [6] shows a theme for quick determination of the focal length of a concave lens. 

High school students usually face difficulty in forming a concept for the magnification and position of a virtal image.  This work will allow the students to verify the lens equation involving the object distance, image distance and the focal length,  using the position and magnification of the virtual image obtained experimentally. After an object is photographed with a smartphone, modern technology enables us to determine the sizes of the images on the camera sensor with micron level  accuracy as has been discussed in [2]. It was shown in that work, that, by photographing an object from two positions, which are separated by more than a minimum distance along the line of sight of the camera, it is  possible to find the width of the object accurately. By knowing the actual width of the object, we find the magnification of the virtual image. From this we determine the focal length of the lens, when the object distance from the lens is known. This focal length has  been checked against the focal length determined by more conventional methods.

\section{Theory}

When an image is formed by a lens, we have the algebraic relation [3,4],
\be\frac {1}{v}-\frac {1}{u}=\frac {1}{f}.\ee Here
$u$ is the object distance and $v$ is the image distance measured from the lens which is assumed to be thin. We will use both a concave and a convex lens for this experiment. By keeping the object distance from a convex lens less than the focal length, we will form a virtual image. $f$ is the focal length of the lens and is positive for a convex lens and negative for a concave lens according to our sign convention. When a virtual image is formed by a convex lens, or by a concave lens, the object and the image  are on the same side of the lens. Transverse magnification produced by the lens is given by \be m=\frac{v}{u}=\frac{I}{O}\ee and is positive for a virtual image. $I$ and $O$ denote the transverse sizes of the image and the object respectively. By photographing the image by a smartphone camera, we will determine $I$, the width of the virtual image formed by the lens. Using equations (1) and (2), we find, \be m=\frac{1}{1+\frac{u}{f}}.\ee Inverting this equation we get \be f=\frac{u}{\frac{1}{m}-1}.\ee By finding the magnification we will find f, the focal length of the lens, using equation (4) when the object distance from the lens is known.

\section {Determination of the width $I$ of the virtual image }

\noindent

In order to find the focal length of the lens we need to know the magnification $m$ produced by the lens.
We first produce a virtual image by a concave lens or a convex lens with the object within its focus distance. The virtual image acts as an object for the camera. We photograph the  object from two positions of the camera; $u_1$ and $u_2$ being  the distances of the camera from the  object (virtual image produced by the lens). In figure 1 we show the configuration of a concave lens along with the camera lens, which has been represented by a thin convex lens. The camera lens in its displaced position has been shown by dashes. $D_1$ and $D_2$ are the distances of the camera from the concave lens. The distance between the camera positions is $D$. 
%From equation (4) we get \be u=f(\frac{1}{m}-1).\ee
If the  transverse magnifications of the  object be  $m_1$ and $m_2$ from the two positions of the camera respectively, we get from equation (4)\be D=f_c(\frac{1}{m_2}-\frac{1}{m_1})\ee where $D= u_2-u_1= D_2-D_1$. Here $f_c$ is the focal length of the camera lens to be distinguished from the focal length $f$ of the lens which produces the virtual image. Our camera is looking at the virtual image through the lens.This displacement $D$ should be along the line of sight of the camera lens. We can write equation (5) as [2]\be I=\frac{\left |D\right |}{f_c\left |(\frac{1}{I_2}-\frac{1}{I_1}\right)|}.\ee We find the width  of the virtual image $I$  by determining $I_1$, $I_2$  and the distance $D$ assuming  we already know the focal length of the camera lens. Here the displacement of the camera can be both towards or away from the thin lens.

\noindent
\section{Experiment}

In figure 2 we show a photograph of our experimental setup. Our object is a ruler which has been fixed to a wall with a two sided glue tape. We put a concave lens in front of the ruler and the smartphone camera looks to the virtual image produced by the concave lens. When the virtual image is well focused by the camera and photographed, a very tiny real image is formed on the camera sensor. This is nothing but a CMOS(CCD) device which stores the images in digital form. Two real images are formed on the camera sensor by taking photographs from two positions. These images are of widths $I_1$ and $I_2$. To find these widths the photographs have to be shared with the Windows or Apple Operating System(OS). The photographs can be opened by the software 'Paint' on Windows (Preview is the software in Apple OS). We choose 5cm of the ruler as our object width as seen in the photograph with the cursor of the Paint software. As soon as we select the width by the cursor, the Paint software gives the width on the sensor in the number of  pixels. The dimension of one pixel in micron can be found from the internet [7] for the particular model number of the camera one is using. By multiplying the number of pixel with the pixel dimension we find the widths $I_1$ and $I_2$. Then we find width $I$ using equation (6). For this calculation we need the focal length of the camera $f_c$. This is also available from the internet [8] for the model number of one's camera lens. We have used our method [9] for finding the focal length of the camera lens. Putting the value of $I$ in equation (4), we find the focal length of the concave lens.

We carry out the same exercise with the convex lens. However, this time we put the object from the convex lens at a distance less than the focal length of the lens and find the focal length.

\section{Experimental Results}

We have worked with a smartphone 'Apple IPhone 12 pro max' to find the focal length  of a concave lens. Results are shown in table 1. The focal length of the camera lens was 0.532cm. One pixel size was $1.7\mu m$. In table 1, $D_1$ represents the distance of the camera from the concave lens and $D$, the displacement of the camera away from the concave lens. The second position of the camera is at a distance $D_2$ from the lens. On sharing the photograph with the Windows OS and then opening it with the 'Paint' software, one can see the enlarged form of the photograph. An object width of 5cm is selected  by the cursor of the software. The software gives the pixel count correspoding to the image width $I_1$ which we call pixel 1 in  tables 1 and 2. Similarly we get pixel 2 as the count from the photograph from the second position of the camera corresponding to the image width $I_2$. By using equation (6) we calculate the  width of the virtual image $I$ produced by the concave lens. From the width  we calculate the magnification $m$. We find the focal length of the concave lens using equation (4) where the ruler was placed at a distance of 8.8 cm from the concave lens. We have taken 10 sets of data varying $D_1$ and $D_2$ keeping the distance between the ruler and the concave lens constant. The average focal length of the concave lens turns out to be $-26.9\pm 0.06$ cm.
 
Exactly the same sequence in the procedure has been followed for the convex lens.  The data is shown in table 2. The smartphone used for the experiment with the convex lens was 'Apple IPhone 12 mini'. The focal length of the camera lens was 0.422cm. One pixel size was $1.4\mu m.$ An object width of 2cm was selected  by the cursor of the Paint software. The distance of the ruler was 9.1cm from the convex lens and this was kept fixed. Here also we take 10 sets of data varying $D_1$ and $D_2$. The average focal length of the convex lens is found to be $17.2\pm 0.04$ cm. We can see that the magnification by the concave lens is less than 1 and that for the convex lens greater than 1 as it should be.

\section{Conclusion}
The advanced technology of modern smartphone cameras has made it possible for the physics students and teachers to perform  a number of physics experiments which were not possible earlier. Here we have used the camera to photograph a virtual image formed by a lens. This virtual image acts as an object and we determine the image size on the sensor of the camera to an accuracy of a few microns. This has allowed us to determine the transverse magnification and position of the virtual image with respect to thin lenses, both concave and convex. From this we determine the focal length of the lenses accurately. Students may find it interesting to make quantitative measurements on the virtual image, not a very common exercise in high school or introductory undergraduate physics laboratories.

\newpage
\begin{table}[ht]
\centering
\caption{Pixel data for the image sizes on the camera sensor from two distances $D_1$ and $D_2$ from the concave lens\\Real Object size = 5.0cm: Object distance from the concave lens= -8.8cm \\Smartphone :Apple IPhone 12 pro max; $f_c=0.532 cm$
; 1 pixel=1.7$\mu m$}
%\begin{ruledtabular}
\begin{tabular}{ccccccccc}
\hline
obs&$D_1$&pixel&$I_1$&$D$&pixel &$I_2$&I&-f\\
no.&cm& 1&cm&cm& 2&cm&cm&cm\\
\hline
1&3.6&1211&0.2059&21.6&376&0.0639&3.76&26.7\\
2&4.5&1110&0.1887&22.4&357&0.0607&3.77&27.0\\
3&5.3&1044&0.1775&8.7&590&0.1003&3.77&27.0\\
4&7.2&894&0.1520&8.5&544&0.0925&3.78&27.3\\
5&5.9&984&0.1673&21.0&357&0.0607&3.76&26.7\\
6&6.3&954&0.1622&16.7&406&0.0690&3.77&27.0\\
7&8.1&834&0.1418&9.6&497&0.0845&3.77&27.0\\
8&11.3&682&0.1159&14.6&370&0.0629&3.77&27.0\\
9&9.7&749&0.1273&11.3&436&0.0741&3.77&27.0\\
10&8.1&834&0.1418&14.9&406&0.0690&3.76&26.7\\

\hline
\end{tabular}
%\end{ruledtabular}
\end{table}

\begin{table}[ht]
\centering
\caption{Pixel data for the image sizes on the camera sensor from two distances of the camera  $D_1$ and $D_2$ from the convex lens\\Real Object size = 2cm: Object distance from the convex lens= -9.1cm\\Smartphone :Apple IPhone 12-Mini; $f_c=0.422 cm$
; 1 pixel=1.4$\mu m$}
%\begin{ruledtabular}
\begin{tabular}{ccccccccc}
\hline
obs&$D_1$&pixel&$I_1$&$D$&pixel &$I_2$&I&f\\
no.&cm& 1&cm&cm& 2&cm&cm&cm\\
\hline
1&12.1&425&0.0595&27.4&222&0.0311&4.23&17.3\\
2&12.1&425&0.0595&43.7&174&0.0244&4.28&17.1\\
3&64.4&156&0.0218&29.5&115&0.0161&4.30&17.0\\
4&93.9&115&0.0161&31.0&90&0.0126&4.26&17.2\\
5&12.1&425&0.0595&52.3&156&0.0218&4.26&17.2\\
6&12.1&425&0.0595&112.8&90&0.0126&4.27&17.1\\
7&64.4&156&0.0218&60.5&90&0.0126&4.28&17.1\\
8&55.8&174&0.0244&69.1&90&0.0126&4.27&17.1\\
9&55.8&174&0.0244&38.1&115&0.0161&4.27&17.1\\
10&55.8&174&0.0244&8.6&156&0.0218&4.17&17.5\\

\hline
\end{tabular}
%\end{ruledtabular}
\end{table}

\begin{figure}[h!]
\centering
\includegraphics[width=12cm]{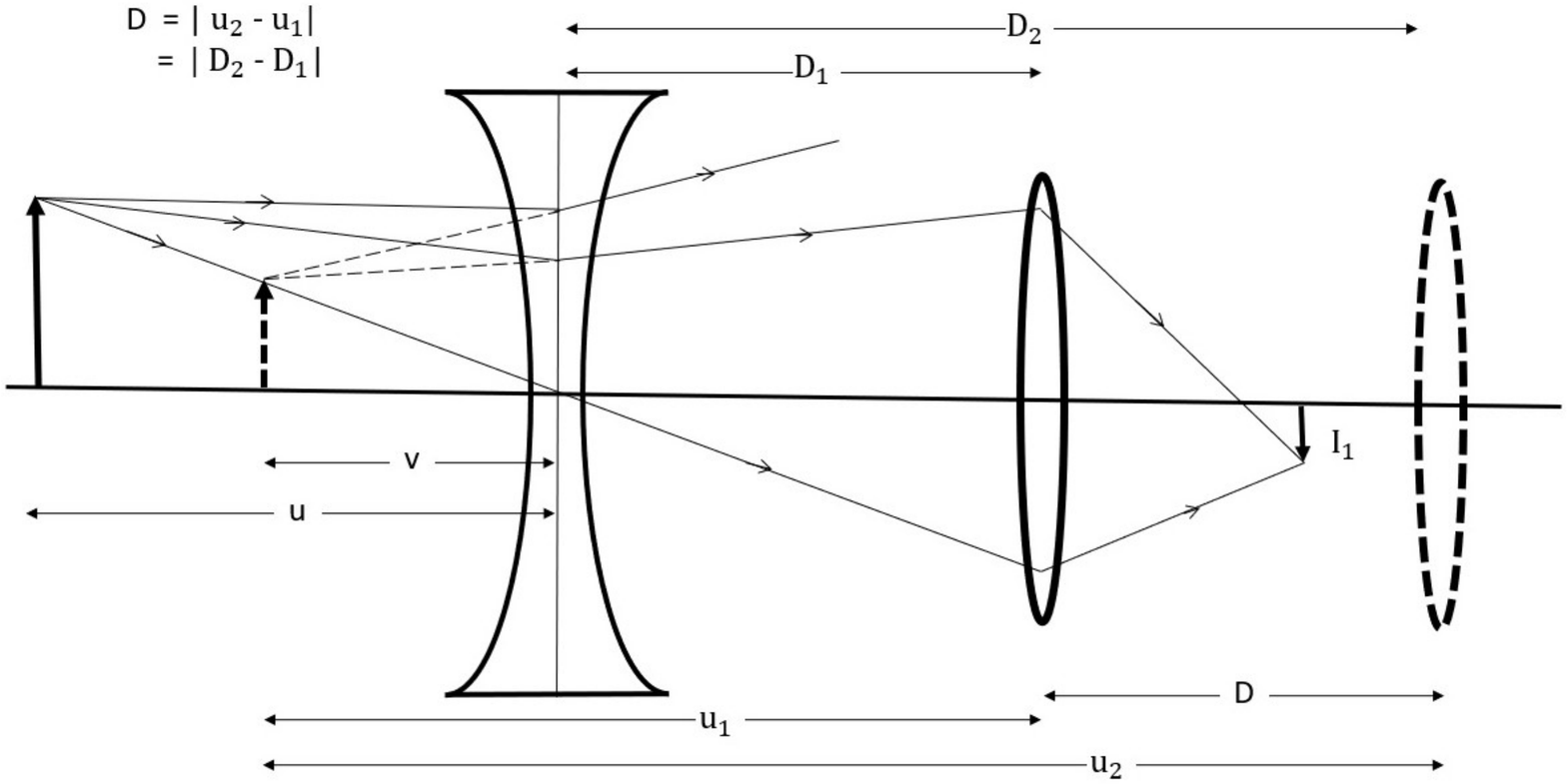}
\caption {Schematic for a virtual image formation by a concave lens and photographing it by a smartphone camera}
\end{figure}

\begin{figure}[h!]
\centering
\includegraphics[width=8cm]{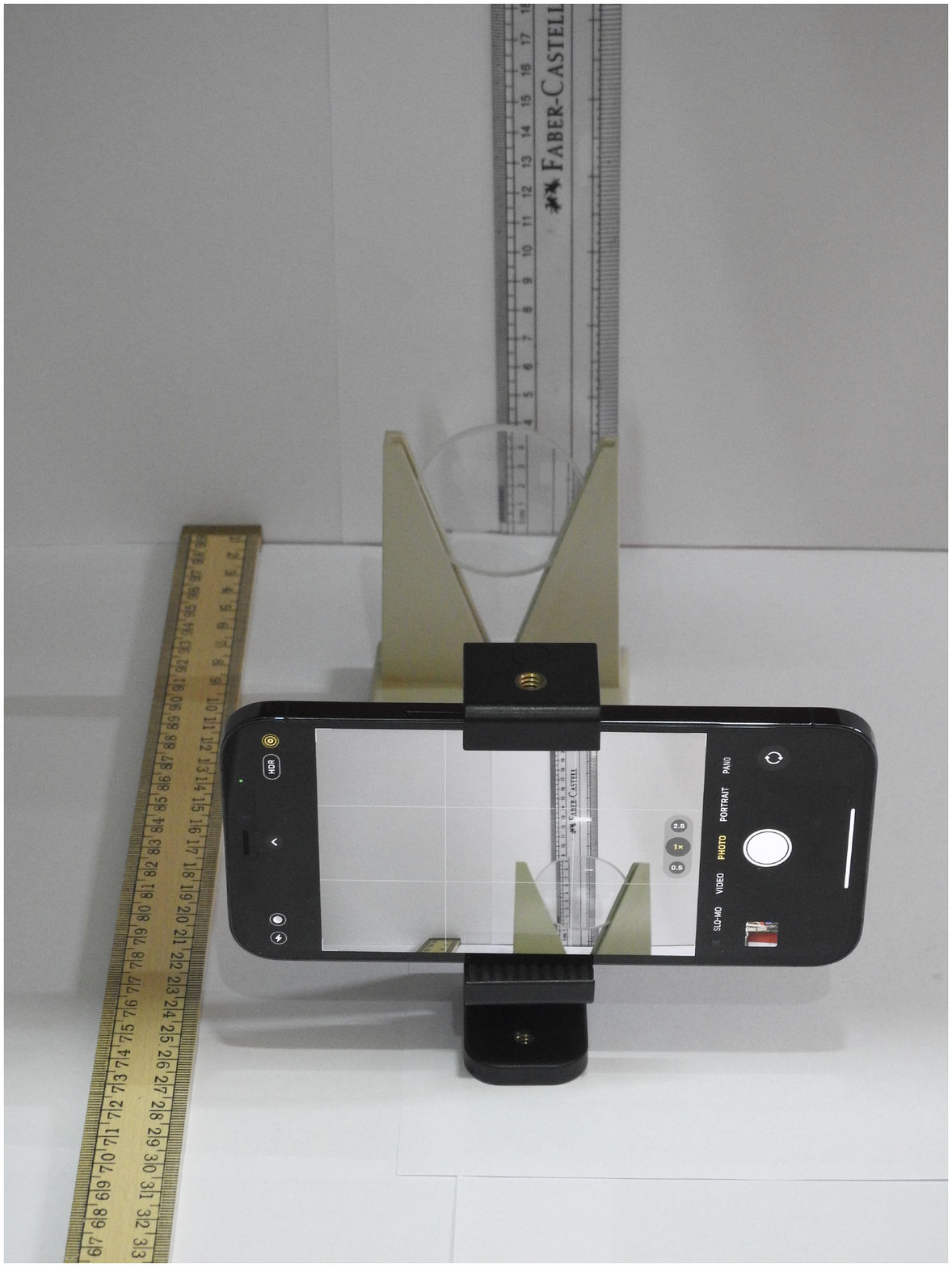}
\caption{Experimental setup with a concave lens}
\end{figure}

\newpage
\begin{center}
References
\end{center}
\begin{enumerate}
\item Monteiro M and Marti A C 2022 Resource letter MDS-1: Mobile devices and sensors for physics teaching Am.J.Phys. \textbf{90}, 328-343.
\item Sarkar S, Pal S K, and Chakrabarti S  Aug 9 2022 Determination of the width and distance of an object with a smartphone  arXiv:2208.05319v1 [physics.ed-ph]. The article has been accepted by The Physics Teacher. After it is published, it will be found at https//doi.org/10.1119/5.0065457.

\item Wang J and Sun W 2019 Measuring the focal length of a camera lens in a smart-phone with a ruler Phys.Teach. \textbf{57} 54 
\item Girot A, Goy N-A, Vilquin A, and Delabre U 2020 Studying ray optics with a smartphone Phys.Teach. \textbf{58} 133
\item Kutzner M D and Snelling S 2016 Measuring magnification of virtual images using digital cameras Phys.Teach. \textbf {54} 503 
\item Dudley S C 1999 How to quickly estimate the focal length of a diverging lens Phys. Teach. \textbf {37} 94
\item https://www.gsmarena.com
\item https://www.metadata2go.com/
\item  Pal S, Sarkar S, Chakrabarti S 2021 11th Nov Determination of the refractive index of water and glass using smartphone cameras by estimating the apparent depth of an object arXiv:2111.06735 v1 [physics.ed-ph].

\end{enumerate}

\end{document}